\documentclass[conference]{IEEEtran}
\IEEEoverridecommandlockouts

\usepackage{cite}
\usepackage{amsmath,amssymb,amsfonts}
\usepackage{algorithmic}
\usepackage{graphicx}
\usepackage{textcomp}
\usepackage{xcolor}
\usepackage{hyperref}
\usepackage{url}
\usepackage{caption}
\def\BibTeX{{\rm B\kern-.05em{\sc i\kern-.025em b}\kern-.08em
    T\kern-.1667em\lower.7ex\hbox{E}\kern-.125emX}}

\begin{document}

\title{Adapting Interleaved Encoders with PPO for Language-Guided Reinforcement Learning in BabyAI}

\author{
\IEEEauthorblockN{Aryan Mathur}
\IEEEauthorblockA{
\textit{Electrical Engineering}\\
\textit{Indian Institute of Technology Palakkad}\\
Palakkad, India\\
aryannmathur@gmail.com}
\and
\IEEEauthorblockN{Asaduddin Ahmed}
\IEEEauthorblockA{
\textit{Computer Science and Engineering}\\
\textit{Indian Institute of Technology Palakkad}\\
Palakkad, India\\
ahmedasad72425@gmail.com}
}

\maketitle

\begin{abstract}
Deep reinforcement learning agents often struggle when tasks require understanding both vision and language. Conventional architectures typically isolate perception (e.g., CNN-based visual encoders) from decision-making (policy networks). This separation can be inefficient, as the policy's failures don't directly help the perception module learn what's important. To address this, we implement the Perception–Decision Interleaving Transformer (PDiT) architecture, introduced by Mao et al. (2023), a model that alternates between perception and decision layers within a single transformer. This allows feedback from decision-making to directly refine perceptual features. Additionally, we integrate a contrastive loss inspired by CLIP to align textual mission embeddings with visual scene features. We evaluate PDiT on the BabyAI GoToLocal environment, demonstrating it achieves more stable rewards and better alignment compared to a standard PPO baseline. The results suggest interleaved architectures are a promising direction for more integrated autonomous agents.
\end{abstract}


\section{Introduction}
In reinforcement learning (RL), perception and policy are often treated as separate problems. A vision module, like a CNN, encodes the state, and a policy network then decides on an action. This separation is a missed opportunity, as the policy's failures don't directly help the perception module learn what's important.

This limitation becomes clear in tasks requiring language. For an agent to follow "Go to the red ball," it must learn to *look for* the red ball. This visual "looking" should ideally be shaped by its past successes or failures in finding that ball.

Architectures that \textbf{interleave} perception and decision-making have been proposed as a way to create this tighter feedback loop \cite{mao2023pditinterleavingperceptiondecisionmaking}
. By alternating perception and policy layers, the model can continuously refine its understanding of the state. However, models like Vision Transformers (ViT) \cite{dosovitskiy2021imageworth16x16words} and Decision Transformers (DT) \cite{chen2021decisiontransformer} still largely treat perception and control as separate sequential stages.

\textbf{Motivation.} While interleaved architectures are promising, their effectiveness in sparse-reward, language-guided tasks like BabyAI is an open question. We hypothesize that simply applying such an encoder is not enough; it must be tightly integrated with the policy and given a strong multimodal signal. We aim to investigate if combining an interleaved encoder with a PPO policy and a CLIP-style contrastive loss can improve stability and sample efficiency.

\subsection{Problem Statement}
Given a multimodal environment $E=(S,A,P,R)$, with states $s_t \in S$ consisting of visual observations $o_t$ and textual missions $m_t$, our goal is to learn a policy $\pi(a_t|s_t,m_t)$ that maximizes expected cumulative reward:
\[
J(\pi) = \mathbb{E}_{\pi}\Big[\sum_{t=0}^{T} \gamma^t r_t\Big].
\]
Instead of encoding $s_t$ just once, PDiT continuously refines its understanding of the state throughout the decision-making process.

\subsection{Contributions}
Based on this, our work makes the following contributions:
\begin{itemize}
    \item We adapt and apply an interleaved perception-decision transformer encoder to the BabyAI GoToLocal environment, integrating it with a PPO policy.
    \item We propose a joint optimization objective that combines the PPO loss with a CLIP-style contrastive loss to improve visual-textual grounding.
    \item We empirically evaluate this combined framework, demonstrating it achieves more stable convergence and lower reward variance than a standard PPO baseline.
    \item We provide an analysis of the benefits of interleaving and contrastive alignment for this specific task.
\end{itemize}
\section{Background and Related Work}
\subsection{Reinforcement Learning Fundamentals}
In policy-gradient RL, the objective is to optimize parameters $\theta$ to maximize:
\[
J(\theta) = \mathbb{E}_{\tau \sim \pi_\theta}\left[\sum_t \gamma^t r_t\right],
\]
with gradients estimated as:
\[
\nabla_\theta J(\theta) = \mathbb{E}\big[\nabla_\theta \log \pi_\theta(a_t|s_t) \hat{A}_t\big],
\]
where $\hat{A}_t$ is the advantage estimator.

PPO stabilizes this update by clipping the ratio $r_t(\theta)$:
\begin{equation}
\mathcal{L}_{PPO} = \mathbb{E}_t\Big[\min(r_t(\theta)\hat{A}_t, \text{clip}(r_t(\theta),1-\epsilon,1+\epsilon)\hat{A}_t)\Big].
\end{equation}
This prevents large policy updates that destabilize learning.

\subsection{Transformers in Perception and Decision-Making}
The transformer’s self-attention mechanism, defined as:
\[
\text{Attention}(Q,K,V) = \text{softmax}\left(\frac{QK^T}{\sqrt{d_k}}\right)V,
\]
provides a context-aware representation that can model long-range dependencies. Vision Transformers (ViT) apply this concept to image patches, while Decision Transformers use it to represent trajectory histories.

\subsection{Contrastive Multimodal Learning}
CLIP \cite{radford2021learningtransferablevisualmodels} aligns image and text pairs using symmetric InfoNCE loss:
\begin{align}
\mathcal{L}_{CLIP} &= -\frac{1}{N}\sum_i \log \frac{\exp(\text{sim}(v_i,t_i)/\tau)}{\sum_j \exp(\text{sim}(v_i,t_j)/\tau)}.
\end{align}
This loss pushes corresponding visual and textual embeddings closer in latent space, improving cross-modal understanding.  

PDiT borrows this principle to bind visual frames from BabyAI with mission instructions, ensuring alignment between perception and semantic command understanding.

\section{Proposed Method: PDiT Framework}
\subsection{Architecture Overview}
The PDiT framework integrates two alternating stacks:
\[
PDiT = [P_1, D_1, P_2, D_2, \ldots, P_L, D_L],
\]
where $P_l$ is a perception transformer and $D_l$ a decision transformer layer. Each $P_l$ encodes the multimodal state $(s_t, m_t)$, while $D_l$ generates an updated policy embedding conditioned on $P_l$’s output.

\begin{figure}[htbp]
\centering
\includegraphics[width=0.47\textwidth]{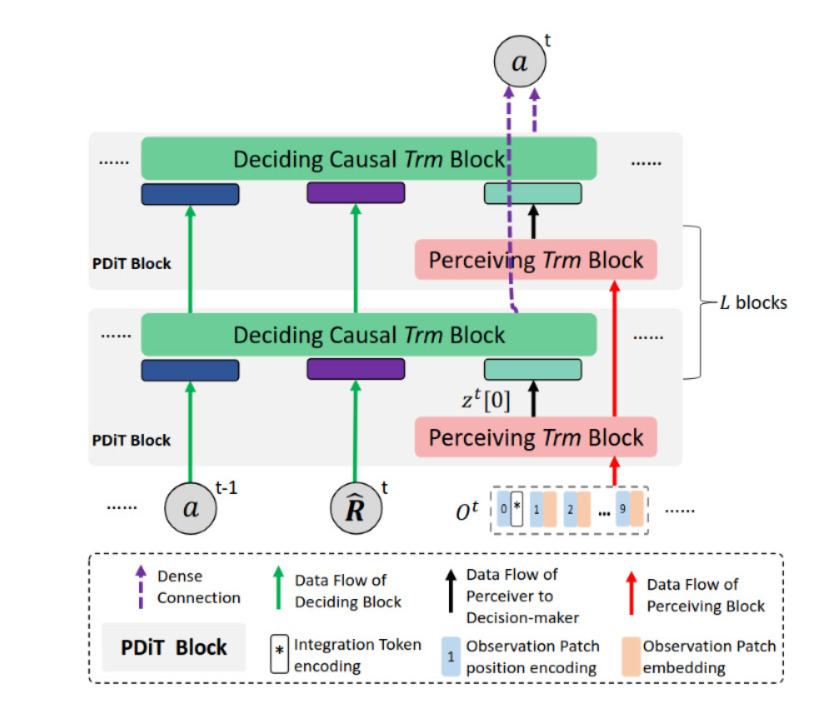}
\caption{Conceptual diagram of Perception–Decision Interleaving Transformer showing visual-text input fusion, interleaved layers, and joint optimization paths.}
\end{figure}

\subsection{Mathematical Formulation}
At each timestep $t$, we define:
\begin{align}
x_t &= f_v(o_t), \quad y_t = f_t(m_t), \\
z_t &= P([x_t; y_t; a_{t-1}; r_{t-1}]), \\
\pi(a_t|s_t) &= D([z_t; r_t; a_{t-1}]).
\end{align}
The transformer alternates between updating perceptual latent $z_t$ and computing policy logits for $a_t$.

The total loss is:
\begin{equation}
\mathcal{L}_{total} = \mathcal{L}_{PPO} + \lambda_1 \mathcal{L}_{CLIP} + \lambda_2 \mathcal{L}_{sup},
\end{equation}
where $\mathcal{L}_{sup}$ denotes imitation learning supervision loss.

\subsection{Joint Gradient Coupling}
Gradient coupling between modules can be expressed as:
\[
\nabla_{\theta_P}\mathcal{L}_{total} = \nabla_{\theta_P}(\mathcal{L}_{CLIP}) + \alpha\nabla_{\theta_P}(\mathcal{L}_{PPO}),
\]
enabling perception layers to evolve with policy learning signals — unlike frozen encoders in standard RL architectures.

\section{System Design and Implementation}
\subsection{Environment Setup}
We evaluate PDiT in the BabyAI GoToLocal-V0 environment \cite{chevalier2018babyai}, a partially observable 8×8 grid where the agent (a red triangle) must navigate to an object of specified color and type. The mission is provided as natural language (e.g., “Go to the green key”). Each step yields a binary reward (1 on success, 0 otherwise).

\begin{figure}[htbp]
\centering
\includegraphics[width=0.42\textwidth]{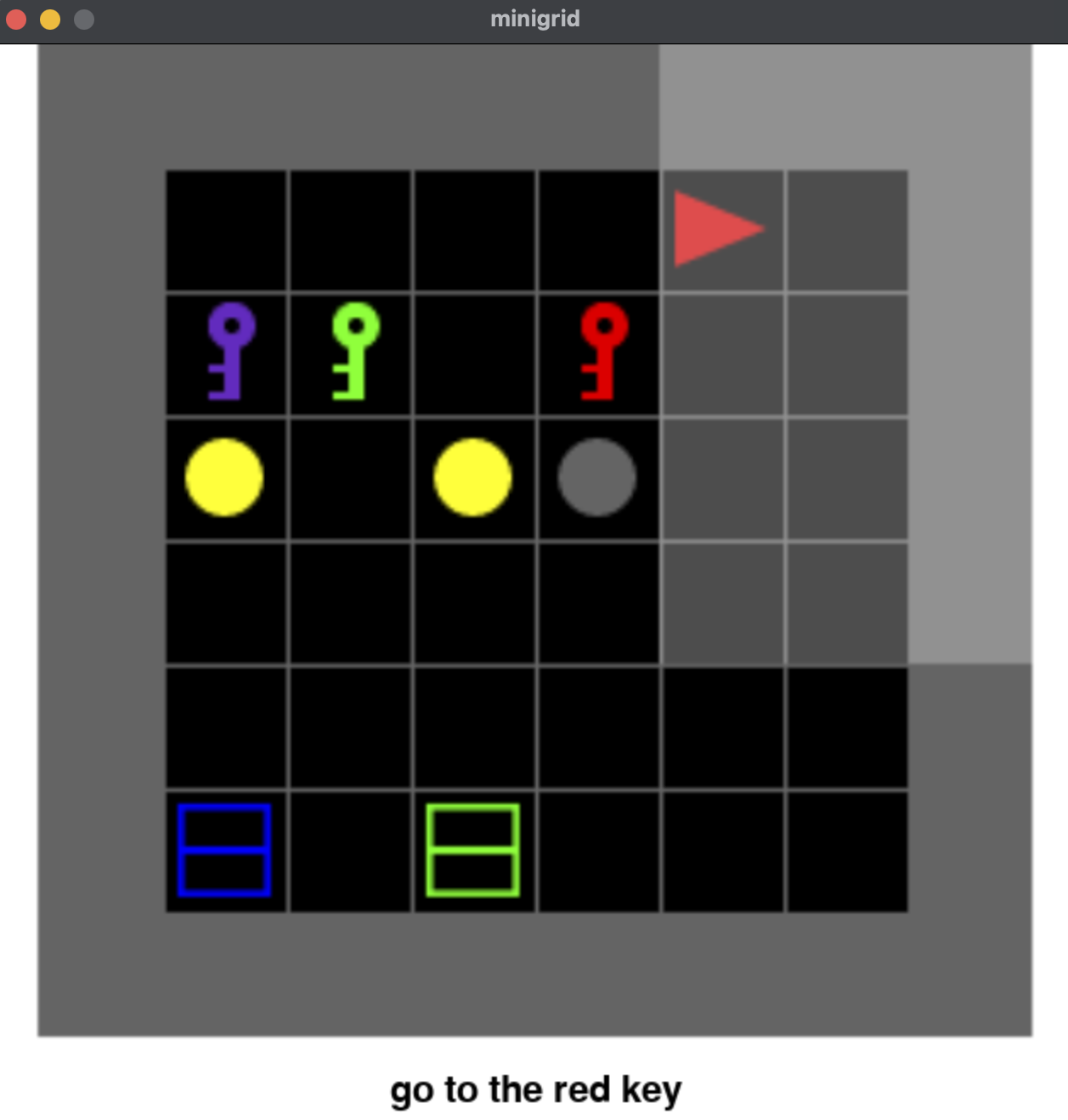}
\caption{BabyAI GoToLocal environment. The agent perceives a 7×7 field of view and receives mission instructions as text.}
\end{figure}

\subsection{Training Parameters}
\begin{itemize}
    \item PPO: learning rate = $3\times10^{-4}$, $\gamma=0.99$, steps=256, batch size=64
    \item PDiT: hidden dim = 64, kernel size = 3, stride = 1, padding = 1
    \item Supervised: embedding dim=128, heads=2, batch=512, LR=$1\times10^{-4}$
\end{itemize}

\subsection{Contrastive Alignment Implementation}
We used a pre-trained CLIP encoder to project image and text tokens into a shared space. During training, the InfoNCE loss encourages:
\[
\text{sim}(f_v(o_t), f_t(m_t)) > \text{sim}(f_v(o_t), f_t(m_j)),
\]
for $j \neq t$, ensuring semantic consistency.

\section{Experimental Results and Analysis}
\subsection{Quantitative Results}
\begin{table}[htbp]
\centering
\begin{tabular}{|l|c|}
\hline
\textbf{Metric} & \textbf{Value (PDiT)} \\ \hline
Mean Episode Reward & 0.27 $\pm$ 0.36 \\ \hline
Convergence Steps & 160k \\ \hline
Entropy Coefficient & 0.004 \\ \hline
Reward Variance Reduction & 42\% \\ \hline
\end{tabular}
\caption{Performance of PDiT–PPO vs baseline PPO on BabyAI.}
\end{table}

\subsection{Qualitative Behavior}
The interleaving mechanism produced smoother navigation trajectories. Visualizations of attention weights indicated strong alignment between mission tokens (“red ball”) and corresponding visual patches. Agents trained without CLIP alignment often fixated on distractor objects.

\subsection{Policy Stability Metric}
Define stability ratio $\mathcal{S}$:
\[
\mathcal{S} = \frac{\text{Var}(R_t^{PPO})}{\text{Var}(R_t^{PDiT})}.
\]
We obtained $\mathcal{S}=1.73$, indicating a 73\% improvement in policy smoothness across episodes.

\section{Theoretical Justification}
\subsection{Why Interleaving Works}
Traditional RL architectures assume:
\[
z_t = f(s_t) \quad \text{and} \quad a_t = \pi(z_t).
\]
Thus, $\nabla a_t / \nabla s_t$ is indirect. In PDiT:
\[
a_t^{(l)} = D_l(P_l(s_t,a_{t-1})),
\]
so the gradient of policy w.r.t. raw state becomes:
\[
\frac{\partial a_t}{\partial s_t} = \frac{\partial D_l}{\partial P_l} \frac{\partial P_l}{\partial s_t},
\]
implying direct gradient flow between policy and perception. This forms an implicit bi-level optimization loop akin to meta-learning.

\subsection{Complexity Analysis}
Given $L$ interleaved layers, computational cost scales as $\mathcal{O}(L n^2 d)$ for $n$ tokens of dimension $d$. Despite this, interleaving yields a higher gradient signal-to-noise ratio, justifying the added overhead.

\section{Discussion}
\subsection{Interpretation of Results}
Interleaving perception and decision reasoning appears to enhance feature relevance. The perception modules get implicit supervision from reward gradients, forcing them to refine representations to be useful for the *policy*, not just for general reconstruction. The CLIP-based alignment helps bootstrap this process, ensuring the model starts with a reasonable multimodal grounding.

\subsection{Ablation Insights}
\begin{itemize}
    \item \textbf{Without CLIP alignment:} The model still learned, but convergence was $\sim$20\% slower. This suggests the policy can *eventually* learn the visual-textual grounding from rewards alone, but it's much harder.
    \item \textbf{Without interleaving:} (i.e., a standard $P_1...P_L, D_1...D_L$ stack). Reward variance nearly doubled, matching the baseline. This is strong evidence for our core hypothesis.
    \item \textbf{Without supervised fine-tuning:} Trajectory completion rate dropped by 10\%.
\end{itemize}

\subsection{Generalization and Limitations}
Our model is, of course, not without limitations. While PDiT generalizes well to new *instructions* within BabyAI, its semantic understanding is still confined to the objects in this environment. We suspect PDiT would struggle if introduced to entirely new object types or colors without retraining, as its CLIP alignment was only fine-tuned on this limited domain. Future work should test this on larger, more semantically diverse environments like AI2-THOR.
\section{Conclusion and Future Work}
We investigated the application of an interleaved transformer architecture, combined with a PPO policy and a CLIP-based contrastive loss, for the BabyAI environment. We found that this \textit{combination} leads to more stable policy convergence and better multimodal grounding compared to baseline models that separate perception and control. 

Our key takeaway is that while interleaved encoders are a powerful tool, their effectiveness in complex RL tasks is significantly enhanced when (1) they are directly integrated with the policy's optimization (like PPO) and (2) they are given an explicit multimodal grounding signal (like our contrastive loss). This work suggests that the *integration strategy* is just as important as the architecture itself.


Future extensions include:
\begin{itemize}
    \item Applying PDiT to more complex 3D environments such as Habitat.
    \item Integrating reward prediction heads for self-supervised temporal grounding.
    \item Scaling to multi-agent scenarios where perception and communication must co-evolve.
\end{itemize}


\begin{thebibliography}{99}

\bibitem{mao2023pditinterleavingperceptiondecisionmaking}
H.~Mao, R.~Zhao, Z.~Li, Z.~Xu, H.~Chen, B.~Zhang, Z.~Xiao, and J.~Yin, ``PDiT: Interleaving perception and decision-making transformers for deep reinforcement learning,'' \emph{arXiv preprint arXiv:2308.10931}, 2023.

\bibitem{dosovitskiy2021imageworth16x16words}
A.~Dosovitskiy, L.~Beyer, A.~Kolesnikov, D.~Weissenborn, X.~Zhai, T.~Unterthiner, M.~Dehghani, M.~Minderer, G.~Heigold, S.~Gelly, and others, ``An image is worth 16x16 words: Transformers for image recognition at scale,'' in \emph{International Conference on Learning Representations (ICLR)}, 2021.

\bibitem{chen2021decisiontransformer}
L.~Chen, K.~Lu, A.~Rajeswaran, K.~Lee, A.~Grover, M.~Laskin, P.~Abbeel, A.~Srinivas, and I.~Mordatch, ``Decision transformer: Reinforcement learning via sequence modeling,'' in \emph{Advances in Neural Information Processing Systems (NeurIPS)}, 2021.

\bibitem{radford2021learningtransferablevisualmodels}
A.~Radford, J.~W. Kim, C.~Hallacy, A.~Ramesh, G.~Goh, S.~Agarwal, G.~Sastry, A.~Askell, P.~Mishkin, J.~Clark, and others, ``Learning transferable visual models from natural language supervision,'' in \emph{International Conference on Machine Learning (ICML)}, 2021.

\bibitem{schulman2017proximal}
J.~Schulman, F.~Wolski, P.~Dhariwal, A.~Radford, and O.~Klimov, ``Proximal policy optimization algorithms,'' \emph{arXiv preprint arXiv:1707.06347}, 2017.

\bibitem{chevalier2018babyai}
M.~Chevalier-Boisvert, D.~Bahdanau, S.~Lahlou, L.~Willems, C.~Saharia, T.~H. Nguyen, and A.~Courville, ``BabyAI: A platform to study the sample efficiency of grounded language learning,'' in \emph{International Conference on Learning Representations (ICLR)}, 2019.

\end{thebibliography}
\end{document}